# STRESS TEST FOR BERT AND DEEP MODELS: PREDICTING WORDS FROM ITALIAN POETRY

Rodolfo Delmonte, Nicolò Busetto

Ca Foscari University, Venice (Italy)

*ABSTRACT*

*In this paper we present a set of experiments carried out with BERT on a number of Italian sentences taken from poetry domain. The experiments are organized on the hypothesis of a very high level of difficulty in predictability at the three levels of linguistic complexity that we intend to monitor: lexical, syntactic and semantic level. To test this hypothesis we ran the Italian version of BERT with 80 sentences - for a total of 900 tokens – mostly extracted from Italian poetry of the first half of last century. Then we alternated canonical and non-canonical versions of the same sentence before processing them with the same DL model. We used then sentences from the newswire domain containing similar syntactic structures. The results show that the DL model is highly sensitive to presence of non-canonical structures. However, DLs are also very sensitive to word frequency and to local non-literal meaning compositional effect. This is also apparent by the preference for predicting function vs content words, collocates vs infrequent word phrases. In the paper, we focused our attention on the use of subword units done by BERT for out of vocabulary words.*

## 1. INTRODUCTION

In this paper we report results of an extremely complex task for BERT: predicting the masked word in sentences extracted from Italian poetry of beginning of last century, using the output of the first projection layer of a Deep Learning model, the raw word embeddings. We decided to work on Italian to highlight its difference from English in an extended number of relevant linguistic properties.

The underlying hypothesis aims at proving the ability of BERT [1] to predict masked words with increasing complex contexts. To verify this hypothesis we selected sentences that exhibit two important features of Italian texts, non-canonicity and presence of words with very low or rare frequency. To better evaluate the impact of these two factors on word predictability we created a word predictability measure which is based on a combination of scoring functions for context and word frequency of (co-)occurrence. The experiment uses BERT assuming that DNNs can be regarded capable of modeling the behaviour of the human brain in predicting a next word given a sentence and text corpus - but see the following section.

It is usually the case that paradigmatic and syntagmatic properties of words in a sentence are tested separately. In this experiment we decided to test them together by combining non-canonicity and infrequent word choice. Italian poetry is characterized by two features, which by the way were also present in Shakespeare's Sonnets: presence of rare and infrequent words, organized into peculiar or non-canonical structures, i.e. structures where the order of constituents was inverted or discontinuous. We carried out the experiment in two phases using a total of 80 sentences and some 1000 tokens. In a first phase we collected 51 sentences extracted from poems for a total number of 700 words and we masked 435 of them. In a second experiment we took the seven sentences most difficult to predict and turned their syntactic structure into canonical order to check the relevance of the unusual syntactic constructions. In a third experiment we chose 11 noncanonical sentences from the VIT treebank taken from newswire domain and did the same: thus we ended up with 29 sentences. We ran BERT both with non-canonical and canonical version to verify the import of syntactic structure.

The most important feature of the experiment is that all sentences are characterized by non-canonical structures. Italian is a language in which non-canonical structures are fairly common due to the weakly configurational nature of the language and to the existence of the pro-drop parameter that allows sentences to freely omit lexically expressed subjects [2]. We then operated on the dataset in two ways: at first we reformulated the text obtained modifying each sentence structure in order to make it canonical. The choice of sentences from poetry has been done in order to focus on the effects of context in conjunction with word level frequency effects - a thorough syntactic and semantic description of these sentences can be found in [3]. The reason for this choice is that poetry is the only domain where rare words are used consistently thus making available a full real context of use for (very) low frequency words. The combined effect of using rare words in a non-canonical syntactic configuration and then restructuring the same sentence with a canonical structure allowed us to make important comparisons.

Non-canonical sentences in Italian can be found in great number due to the pro-drop nature of the language which thus resembles Chinese and Japanese [4]. In addition, Italian is a morphologically rich language thus possessing a very large vocabulary of unique wordforms which, if compared to the total number of wordforms obtainable from the WordNet list of citation forms for English is an order of magnitude higher – from 500K to 5 million wordforms in Italian, only considering the corresponding number of grammatical categories [5]. We already discussed elsewhere [6] that languages like Italian, which have a rich morphology, need embeddings with higher dimensions and a vocabulary size more than doubled in order to account for the variety of semantically relevant wordforms.

When referring to context in BERT, the whole preceding sentence portion is included. BERT being bidirectional the context will apply to both the right and the left previous sequence of tokens. However, when referred to Distributional Semantic Models, the context is usually determined by the number (2 to 5) of co-occurring tokens to be considered when building vectors for word embed- dings: if the masked word is the first word in the sentence only the right context will be available and this fact reduces the ability of prediction as shown by our data. The result of our experiment shows that DNNs are very sensitive to context and that frequency of occurrence is less relevant for word predictability – but see below.

The paper is organized as follows: in the following section, we introduce briefly state of the art on the problem of word predictability as seen from the linguistic and cognitive point of view; in section three we deepen the question of non-canonicity and its relevance for complexity evaluation; in section four we present the experimental setup and the typology of non-canonical structures contained in our dataset; section 5 presents experimental results and discuss the import of context for word predictability, then our conclusion. In the Appendix we reported the translated version of the seven most difficult sentences together with the 11 sentences taken from newswire domain, while the detailed analysis is contained in the Supplemental Material.

## 2. Word Predictability in Cognitive and Psycholinguistic Research

Word prediction or predictive language processing has been a foundational topic for psycholinguistic research in the last 50 years or so for all that concerns human sentence processing and comprehension. In this paper we intend to exploit the hypothesis presented lately in a number of papers [7, 8] where human word predictivity is compared and tested by the performance of DNNs in next-word-prediction tasks. In particular, in their conclusion, Schrimpf et al. comment on the results of their findings defining them as an attempt to create a viable hypothesis for modeling predictive language processing in human brain by the use of predictive artificial neural networks, specifying that so-called “transformer” models - BERT - are best-performing models. In another paper (see [9]), they had already come to the conclusion that it is by the use of working memory as a whole that word predictivity works: i.e. the integration of all levels of language processing, lexico-semantic, syntax and knowledge of the world conspire to make word prediction viable in order to carry out the primary function of human language, “the extraction of meaning from spoken, written or signed words and sentences (see [8:2]).

The question of word frequency and their predictability is dealt with in great detail in a paper by [10]. Words which have high predictability scores are also those which are somehow more related to the prior context, and words which are more related to the prior context are also easier to integrate semantically. “...there is no such thing as an unexpected word; there are only words which are more or less expected." (ibid. 309). In this approach, predictability changes from one word to the next due to syntactic and semantic constraints, eventually coming to the conclusion that speakers tend to choose words more likely to occur in a given context.

Estimating the level of difficulty or the “surprisal” or unpredictability - of a word in a given context is done by the negative log probability measure which counts as 1, words fully predictable and as 0 those unpredictable, where the former ones convey no additional information as opposed to the latter. Thus, in a serial-search model imagining lexical access in a frequency sorted lexicon, the 100th most frequent word would take twice as long to access as the 50th most frequent word. As a consequence, most frequent words are less informative and are easier to pronounce and to understand. However, this may only be regarded as a theoretically viable hypothesis since even when words are infrequent and unknown they may still serve to formulate some meaning related bit of information and help in understanding the content of the utterance.
From the results obtained in our experiment based on BERT raw embeddings, both frequency and context conjure to establish word predictability. In some cases it is clearly the low frequency to prevent embeddings to be made available, but in other cases - see the example of the ambiguous word "ora"/now-hour below - even though the word and the local context is fairly typical, the word is not predicted.

A partly similar approach has been attempted by Pedinotti et al.[11], in a paper where they explore the ability of Transformer Models to predict transitive verb complements in typical predicate-argument contexts. Their results show clearly the inability to predict low frequency near synonyms, thus confirming the sensitivity of BERT-like models to frequency values. The experiment also included a version of the dataset where the surface syntactic structure of the sentences was modified in order to introduce non-canonical structures. In fact this was only limited, though, to two cases: interrogative and cleft-structures. The second structure showed how the model suffered from non-recurrent word order by an important drop in performance (from 70 to 38% accuracy).

Another parameter which has loomed large in the cognitive literature is the relevance of the effort/time required to pronounce/read a word: a short word, both phonetically and as grapheme, is preferred and confirmed in an experiment based on semantic grounds by Mahowald et al. [12], where pairs of near synonym words inserted in frame sentences and user have consistently chosen the shortest ones as the most predictable. This seems to be confirmed by the well-known fact that the top range of frequency lists of wordforms are occupied by short words thus confirming the inverse correlation existing between word length and frequency. Most frequent words are not only the shortest but the ones with more senses as confirmed in a paper by Piantadosi et al. [13], hence the more frequent. To verify this we inspected the top 200 words in the frequency lists of ItWac for Italian and English and counted their number of syllables with the following results: Italian has 75 monosyllabic words and 125 words with more than one syllable; English has 149 monosyllabic words and 51 words with more syllables. The two languages have an opposite distribution as has also been documented in a previous paper [4]. In addition, English top 200 words contain only 30 content words, while Italian contains 61 content words, ten of which are morphological variants, English has only one morphological variant.

## 3. Unusual or Non-Canonical Syntactic Constructions and Cognitive Models

First of all we will assume that Italian is a configurational language with a neutral or canonical sentence structure of type SVO (Subject-Verb-Object). In fact Italian can also be regarded as a weak non-configurational language thanks to the presence of some important typological parameters: free subject inversion, pro-drop and non-lexical expletives. It also has no wh- in situ, no preposition stranding, no deletable complementizers, no impersonal passives, no parasitic gaps with the same argument.

In cognitive linguistic terms, we are now referring to what is encoded at the informational structure level by means of surface or syntactic constituency and word order. However, we assume that functional or relational interpretation of syntactic structures is necessary together with semantic processing in order to understand what semantic roles are associated with displaced grammatical functions. Thus, whenever a subject is found in inverted position, we will be talking of a pragmatically motivated displacement producing a focus on an otherwise topic-related function. Subjects are always understood as topic, i.e., as referring "what the sentence is about," and constituting old information, and in that role they come before the verb. This is obviously subject to the category of word order a given language obeys to, and Italian - and also English for that matter - is an SVO language. On the contrary, a focus is defined as "the essential piece of new information carried out by a sentence" and is usually placed after the verb where the "comment" portion of the sentence is. Or, if we consider the subdivision of the sentence into theme/rheme, the subject would be the theme, while the comment will become the rheme.

We take complexity measures to be sensible to noncanonical structures (hence NCS) that are pragmatically motivated and are used to encode structured meaning with high informational content, related to the FOCUS/TOPIC nonargument functions in LFG (see Bresnan, [14-15]). Noncanonical structures can be said to help the reader or interlocutor to better grasp the intended (pragmatically) relevant meaning in the context of use (see Birner and Ward [16-17]). In [18], the authors "report an investigation into the online processing of non-projective dependencies in the context of prominent contemporary theories of syntactic comprehension" (ibid.:3) which is totally dedicated to extraposed relative clauses in order to show that readers develop a high level of expectancies for the presence of a possible non-projective or noncanonical modifying structure of an already computed NP head.

Predictability of a certain NCS highly depends on its frequency of use in given contexts which in turn is strictly related to the possibility of a “surprisal” effect in the reader. Frequency determines probability of usage, and it is intuitively easy to surmise that whenever linguistic items and structures are less probable, they will be more surprising [19-22]. In order to show the impact of NCS on the parameter of novelty and unusualness, we will compare English syntactic structures to Italian ones. Italian NCS are relatively highly represented in text, as the following table shows, with quantities extracted from VIT, our treebank (see [2]), where they have been explicitly marked with the labels indicated below.

The final percentage is computed on the total number of constituents in written texts, amounting to 230,629. If we compare these data with those made available for Latin (see [23]), where the same index amounts to 6.65% – data taken from the Latin Dependency Treebank containing some 55,000 tokens – we can see that Italian and Latin are indeed very close. The second percentage is computed by dividing up number of NCS/total number of simple sentences. As for tree projectivity in English, we refer to the Penn Treebank (here marked as PT), where numbers are fairly low as can be seen in the following Table 2.

The total number of constituents for PT amounts to 720,086. Percent of NCS are computed on the number of total utterances, while percentage of unexpressed subjects are computed on the number of total simple sentences. The nonprojectivity index for PT would then amount to 0.01004%. Expectancies for an Italian speaker for presence of a NCS are thus predictable to be fairly high, due to processing difficulties raised by number of unexpressed subjects. This will not apply to an English speaker because NCS are infrequent and used only in specific contexts and situations.

### 3.1. The Dataset and the State-Of-The-Art

As said above, Italian is very rich in number and types of NCS. This is also due to its being a direct derivation from Latin, a free word-order language (see [4]). Our approach has been previously adopted by other researchers but with slightly different aims that we describe in what follows. The first work is by Paccosi et al. [24] where the authors present a new dataset of Italian based on "marked" sentences, which is then used to verify the performance of a neural parser of Italian (TINT) on the dataset. The result for LAS dependency structures is 77%, 3 points below the best results previously obtained on the UD corpus of Italian, which was 80% accuracy. This result confirms previous work documented also in [25] with a small dataset containing strongly marked sentences, which have been included in the text used in this paper, where the results were well below 50% accuracy. The authors make a detailed description of the type of marked structures they annotated in their treebank corpus. It is a list of seven structures - cleft, left dislocated, right dislocated, presentative "ci", inverted subject, pseudo-clefts, hanging topic - with a majority of Cleft sentences and Left dislocated sentences. As said above, similar results are obtained by the experiment presented in the paper by Pedinotti et al. [11] where in Section IV they test the ability of Transformers - they use RoBERTa - on a small dataset with surface syntactic structures different from the recurrent word order. They modify the sentences to produce cleft and interrogative versions of the same sentences. The result for core semantic roles - this is what they are testing - is a dramatic drop of performance from 0.65 of correlation in canonical transitive versions down below 0.35.

When compared to the corpuses above, our dataset is smaller but it contains many more types of marked constructions, which makes it more difficult to come to terms with, and this is due mainly to presence of sentences from the poetry domain. Focusing only on the third experiment and its dataset, we present now the noncanonical structures contained in this dataset, including 7 sentences from the poetry domain and 11 sentences taken from the VIT treebank from the bureaucratic and newswire domains that can be found with detailed comments in the Appendix. Similar structures can be found in the wider dataset of the first experiment.

- ***complete argument inversion*** (the complement is fronted and the subject is in post verbal position) in sentence 7B - with copula deletion, and in sentence 17B with infinitival structure as subject;
- ***object fronting*** (the object comes before the subject at the beginning of the sentence) in sentence 2A and 5A;
- ***adjective extraction*** (the adjective is extracted and fronted from the noun phrase) in sentence 13A and 14A;
- ***PPadjunct preposing from participial clause*** in sentence 1B and 13A;
- ***lexical verb left extraction*** (the main verb - untensed non-finite - is positioned before the auxil- iary/modal) in sentence 3A;
- ***subject right dislocation*** (the subject is positioned after the complements) in sentence 3A and 6B; ***subject and object fronting*** (the subject comes before the object and both are positioned before the main verb) in sentence 4A and 5A; *PPspecification extraction from the noun phrase and fronted to the left* in sentence 5A;
- ***clitic left dislocation*** in sentence 8B;
- ***object right dislocation*** (the object is positioned after the indirect object or the adjuncts) in sentence 10B;
- ***parenthetical insertion*** (a parenthetical is inserted after the subject before the main verb) in sentence 11B and 16B;
- ***adjective right extraction*** (the adjective is extracted from the noun phrase and positioned after the noun adjuncts) in sentence 11B and 14A; *PPspecification right stranding* - the PPof is stranded to the right out of the noun phrase in sentence 14B;
- ***lexical verb right extraction*** (the main verb - un- tensed non-finite - is positioned after the complements) in sentence 12A;
- ***double parenthetical insertions*** (after the subject and after the verb complex and before the complements) in sentence 15B and 16B;
- ***clitic left dislocation with subject fronted as hanging topic* in sentence 18B.**

## 4. THE EXPERIMENTAL SETUP

We assume that word predictability can be characterized by two parameters: word (co-occurrence) frequency/ies and linguistic complexity measured by syntactic/semantic related scoring functions. We evaluate word co-occurrence frequencies by means of embeddings as the cosine value made available by BERT in its first projection layer, using pre-trained models and no fine-tuning. We produced the whole experiment leveraging the ability of the Huggingface implementation [27]. We used BERT – with the Italian model taken from UWAC corpus, Umberto-commoncrawl - and examined the output of the first or projection layer[3]. In this way we intended to check the predicting ability of BERT on the masked word, by selecting in turn one content word at a time allowing BERT to use the rest of the sentence as a context to make appropriate predictions. Of course, we are aware of the fact that by training a DNN, its error rate may be reduced in cycles through back propagation. This involves comparing its predicted function value to the training data that we did not intend to use. Error reduction is done by computing the gradient of a cross entropy loss error function and proceeding by specified increments of the weights to an estimated optimal level, determined by stochastic gradient descent, which in the case of a test set, does not necessarily correspond to what has been learnt.
It is a fact that words are represented in a DNN by vectors of real numbers. Each element of the vector expresses a distributional feature of the word - in our case by cosine values. These features are the dimensions of the vectors, and they encode their co-occurrence patterns with other words in a training corpus. Word embeddings are generally compressed into low dimensional vectors (200-300 dimensions) that express similarity and proximity relations among the words in the vocabulary of a DNN model.

In order to evaluate frequency values associated to each masked word, we cleaned the frequency list of Italian wordforms compiled on the basis of ItWaC - which contains approximately 388,000 documents from 1,067 different websites, for a total of about 250M tokens. All documents contained in the PAISA' corpus date back to Sept./Oct. 2010. The itWaC corpus is available at https://wacky.sslmit.unibo.it/ accessed on October, 2021 -, deleting all numbers and websites. Thenwe created a list of 50000 most frequent wordforms to be used to check what words would be included by a model created on the basis of BERT tokenization module. Wordforms included are up to a frequency value of 1377. The remaining list is cut at frequency value 4, thus leaving out Rare words, made up of Trislegomena, Dislegomena and Hapaxlegomena, which is by far the longest list: it counts 1,642,949 entries. The inclusive List – the list that includes the 50000 plus the rest of wordforms down to and including words with frequency 4, is made up of 513,427 entries. Then, we divided the 50000 vocabulary into two halves: first half with "high" frequency words, including three segments - highest, high and middle frequency words down to 10000 -, second half from 10000 to 1377 we call "low" frequency words. We then consider as "very-low" frequency words those included in the so-called inclusive List - from 1377 down to 4 occurrences -, and the remaining long tail are classified simply as "Rare Words". The final classification is then organized into four classes: High, Low, Very Low and Rare. To make frequencies more visible, we mark with one asterisk words belonging to "Low", with two asterisks words belonging to "Very-Low", and three asterisks "Rare" words.

## 5. Experimental Results and Discussion

In the experiment we ran BERT by masking each content word and some function word, one at a time in order to be able to make a detailed error analysis and parameter evaluation. As said above, we carried out three separate experiments to test BERT ability to predict the masked word: in the first experiment we collected 51 sentences extracted from poems for a total number of 700 words and we masked 435 of them. In a second experiment we took the seven sentences most difficult to predict and turned their syntactic structure into canonical order to check the relevance of the unusual syntactic construction. In a third experiment we chose 11 noncanonical sentences from the VIT treebank taken from newswire domain and did the same: we ran BERT both with non-canonical and canonical version to verify the import of syntactic structure. Finally we ran BERT in ML mode fine-tuning the model with VIT corpus that contains 30% of NCS, and used the 51 sentences of poetry as test text. As shown in Table 1. below, accuracy in word prediction considering the first 10 candidates is 29.425%.

Table 1. General Data of the Experiment and Accuracy

| | Tot.No. Words | Masked Ws. | No.Sentences | Correctly Rec. | Accuracy |
|---|---|---|---|---|---|
| PoetryCorpus | 700 | 435 | 51 | 128 | 29.43% |

The evaluation has been carried out on the basis of cosine values, as they were made available by the first ten candidates computed by BERT. Word predictability has been measured by BERT raw word embeddings and their cosine value, by masking one content word at a time - and a few function words. In order to evaluate a sentence level predictability score, we took the cosine value of the recognized masked word disregarding its position; when it was not correctly predicted in the first ten candidates we selected the first candidate and its cosine value. We then added all the values found at sentence level. To evaluate the effect of frequency ranking, each content word has then been searched in the frequency list made available by the ItWac frequency list. Frequency contribution was computed simply by each word position in the frequency list, dividing very frequent words from low frequency ones.

Of the 128 predicted words over 435, 41 were grammatical or function words over a total of 69 in the whole corpus, which were also associated with the highest cosine score. 69 words were predicted in the first three slots of the ten candidates and of course they included 29 function words.

Table 2. Typology of Recognized Masked Words

| | No. Recog.Words | Recog. Gram.Words | Recognized in first 3 slots | Gram. Words in first 3 slots | Ratio of Semantic Ws. |
|---|---|---|---|---|---|
| Recog. Words Percentage | 128 | 41 32% | 69 53.9% | 29 70.73% | 128-41=87 67.97% |

## 5.1. Linguistic Features Associated with the Prediction Task

Before presenting in detail positive results obtained by BERT we will briefly try to motivate the reasons behind the high number of failures in recognition of the masked words. Overall, BERT recognized 128 over 434 words approximately 30% of the masked words. The remaining 306 words failed by BERT may be at first characterized as follows:

28 are OOVWs (out of vocabulary words) and will be analysed in detail in the section below;
the remaining 278 words can be further subdivided into three main subclasses, on the basis of linguistic features associated to the first ten candidates we extracted, as follows:

1. candidates with the same grammatical category of the masked word
2. candidates with a different grammatical category from the masked word
3. candidates with a majority of identical category vs. candidates with a majority of different category

The class of candidates with the same grammatical category and a majority of it is by far the most frequent: 181 words against 97 for categories: different and most different.It is important to note that BERT has been able to characterize the masked word from the point of view of its grammatical category, thus showing a complete control of the linguistic context in terms of its syntax,

If we look at the CONTEXT in which the failed words are located we may note some recurrent features. In particular, in category 1. most words are semantically associated with the masked word: some are quasi-synonyms, some others are quasi-antonyms with opposite meaning. None of this happens with words in category 2. Most words belonging to category 2. where BERT did not manage to guess the same grammatical category, are words with apocope, i.e. words where the final vowel has been elided. None of these cases are present in the first category, where BERT managed to guess the same grammatical class. The remaining cases in category 2 are either due to the presence of a word which has not been recognized or a noncanonical position: we may find "inverted subject NP" position, "object NP preposing", for nouns; "extracted adjective" from a PPof, and "preposed Adjective complement(ACOMP)" for adjectives; and inverted position for possessives. Adverbs have a high freedom in assuming positions at syntactic level so they are not affected, except for frequency time adverb "sempre"/always when position far from the verb it belongs to semantically speacking. There are a few archaic word: "guardo" for "sguardo"/gaze, "verno" for "inverno"/winter, and "meco" an amalgamated clitic+preposition pair no longer used which stands for "con me"/with me.

## 5.2. Subword Units

BERT like other similar Deep Learning models copes with out of vocabulary words (hence OOVW) by substituting it with so-called “subword units”, i.e. sequences of characters selected on the basis of their frequency. As we already remarked in a previous paper (see [6]), this choice destroys the semantics of the masked word and as a consequence the semantics of the sentence under analysis is lost. In particular, we found that BERT showed consistent disruptive behaviour in 34 cases: it treated as unknown <unk> 4 words and complained the fact that the separator symbol <s> has not been used as sentence start in two additional cases. Then it introduced subword units or just punctuation marks in 28 cases, with the result that illegal or inexistent Italian words were generated. It is important to remind that when subword units are generated they erase the masked word, they act as suffixes and attach to the previous word rather than producing a new substituting word.

We analyse here below the list of bad mistakes subdividing them according to the type of behaviour of BERT resulting in different outputs subdivided into 6 subcases, a-f.

Table 3.Typology of Subword Units Generated by *BERT*

| |
|---|
| a. punctuation generated consistently + <unk> |
| b. punctuation generated consistently + <s> and <s>NOTUSED |
| c. subword units generated producing non-words |
| d. subword units generated producing non-words + substitution words |
| e. subword units generated producing non-words + legal Italian words |
| f. subword units generated producing non-words + legal Italian words + substitution words |

As can be easily noticed from the list below, the most frequent case is f. with 13 cases followed by e. with 5 cases.

In the analysis below, we report the masked word, the preceding word to which subword units attach, the list of suffixes, then we list the resulting inexistent words which constitute illegal Italian words after two slashes. After BUT--- are listed legal Italian words for the masked one. Finally, FULL--- is used to list words substituting incorrectly the masked word.

***Case a. : punctuation generated consistently + <ukn>***
"mostri"/show preceded by "petto"/bosom and followed by fullstop
'.', '.', ',', '...', '!', '...', 'è'\is, <unk>, ',', ',.'

"saluto"/greet preceded by "tetti"/roofs and followed by fullstop
'.', ',', '.', '...', '!', ';', <unk>, '...', 'miei'/mine, ': '

"Ulisse"/Ulysses preceded by "Itaca"/Itacha followed by fullstop
'.', '.', <ukn>, ',', '?', '...', '!', '»', '...', '».'

"esclude"/hampers preceded by "guardo"/gaze-archaic+look_1pers_sing_pres_indic and followed by fullstop
'.', <unk>, ',', '...', "intorno"/around, "io"/I, "ancora"/still+anchor, "fisso"/fixed, '!', '.'

***Case b. : punctuation generated consistently + <s> and <s>NOTUSED***
"Così"/So first word followed by "tra"/amid
<s>, <s>NOTUSED, '.', "leggere"/reading+light_plur_fem, ',', "de"/of, <s>NOTUSED, "animo"/animate+soul, "dire"/say, ':'

"Tu"/You first word followed by "non"/not

<s>, <s>NOTUSED, '.', ',', "dire"/say, <s>NOTUSED, "che"/that, "anima"/animates+soul, "leggere"/reading+light_plur_fem, "tempo"/time

***Case c.: subword units generated producing non-words***

"opra"/deeds preceded by "l'agil"/the agile and followed by "de"/of
SUBWORD UNITS = da, dio, era, de, do, dia, fa, lio, ria, enza // l'agilda, l'agildio, l'agilera, l'agilde, l'agildo, l'agildia, l'agilfa, l'agillio, l'agilria, l'agilenza

"silenzi"/silences preceded by "sovrumani"/superhuman and followed by "e"/and
SUBWORD UNITS = issima, tà, ta, issima, ca, ssimo, osa, ca // sovrumanica sovrumaniosa sovrumaniissima sovrumanieta' sovrumanica sovrumanita

***Case d. : subword units generated producing non-words + substitution words***

"canto"/song preceded by "il"/the followed by "avrai"/will have
SUBWORD UNITS = sospi, vesti, mor, sepol, rapi // mor, sepol, vesti, rapi, sospi
FULL--- nome, cor, primo, dolce, corpo

"cantò"/sang preceded by "acque"/waters followed by "fatali"/fatal
SUBWORD UNITS = dì, gge, dette, ggiava, ggia, // acquedì, acquegge, acquedette, acqueggiava, acqueggia, FULL--- ebbe, rende, più, di, rese

***Case e. : subword units generated producing non-words + legal Italian words***

"bove"/ox preceded by "pio"/pious followed by ','+"e"/and
SUBWORD UNITS = vano, vio, vere, ppi, va, vra, vi, ve, ppo, vo // piovo, piovio, piovi, piovano, BUT--- pioppo, piove, piovra, piova, pioppi, piovere,

"esili"/slim preceded by an apostrophe and "com"/like followed by "pensieri"/thoughts
SUBWORD UNITS = i, i, ei, e, è, altri, ai, a, un, hanno // com'i, com'i, com'ei, com'ai, com'e, com'a, BUT--- com'è, com'altri, com'un, com'hanno

"cantando"/singing preceded by "man"/hand with final vowel "o" elided followed by ','+"fassi"/is made
SUBWORD UNITS = cia, isa, tellina, cina, ina, na, ia, mano, data, etta // manisa, BUT--- manetta, mandata, man mano, mania, manna, manina, mancina, mantellina, mancia

"come"/as preceded by "sta"/stay followed by "d'autunno"/of autumn
SUBWORD UNITS = llino, gano, no, cchino, ranno, gnano, cchi, glia, ccano, gliano // stallino, stagano, stano, BUT--- stacchino, staranno, stagnano, stacchi, staglia, staccano, stagliano

"eseguire"/accomplishing preceded by "l'"/the followed by "opere"/deeds
SUBWORD UNITS = ele, inte, uste, oro, amore, anime, ora, onde, oste, altre
l'ele, l'inte, l'uste --- BUT l'oro, l'amore, l'anime, l'ora, l'onde, l'oste, l'altre

***Case f. : subword units generated producing non-words + legal Italian words + substitution words***

"pensieri"/thoughts preceded by "esili"/slender followed by "nel"/in_the
SUBWORD UNITS = scono, ano, vano, tano, rono, i, issimo, ',' // com'esiliscono, com'esilivano, com'esilitano, com'esilirono, com'esilii, BUT--- com'esiliano, com'esilissimo, com'esili',', FULL-- e, che

"sovrumani"/superhuman preceded by "e" followed by "silenzi"/silences
SUBWORD UNITS = terni, mpi, ternità // BUT--- eterni, empi, eternità FULL--- quei, altri, di, dai, là, da, più

"conversare"/talking preceded by "a"/to followed by "l'anima"/the soul
SUBWORD UNITS = erare, rare, lare, scere, mare, urora, vare, gnare // ascere, agnare BUT--- aerare, arare, alare, amare, aurora, avare, FULL--- vivere, riposare,

"può"/can preceded by "riprendere"/recover followed by "a"/to
SUBWORD UNITS = sse, mo, sterebbe, erebbe, sti, mmo, i, bbero, bbe, // riprenderesterebbe, riprendereerebbe, riprenderesse, BUT--- riprenderemo, riprenderesti, riprenderemmo, riprenderei, riprenderebbero, riprenderebbe FULL--- sarebbe

"mezzo"/means preceded by "in"/in followed by "a"/to
SUBWORD UNITS = grati, erti, tenti, sistenti, dotti, tente, dotte // BUT--- ingrati, inerti, intenti, insistenti, indotti, intente, indotte FULL--- corsa

"primavere"/spring preceded by "Se"/If followed by "il"/the
SUBWORD UNITS = condo, mmai, BUT--- Secondo, Semmai, FULL--- fossi, avessi, fosse

"tenerella"/sweet+alteration preceded by "o"/oh, "perivi"/perished followed by fullstop
SUBWORD UNITS = rante, ziosa, blio, de, ppio // orante, BUT--- oziosa, oblio, ohimè, ode, oppio FULL---morte, quasi, donna, morta,

"combattuta"/harsh preceded by "morbo"/illness followed by "e"/and
SUBWORD UNITS = sa, rata, ',', lata, tta, ttata, si, sità, so, //
morbosa, morborata, morbo',', morbolata, morbotta, morbottata, BUT--- morbosi, morbosità, morboso, FULL--- morta

"lasciando"/leaving preceded by "talor"/sometimes with final "a" elided followed by "e"/and
SUBWORD UNITS = ',', ni, ini, ghi, dici, mi, nali, menti // ',', talorni, talorini, talorghi, talordici, talormi, talornali, talormenti FULL--- libri, fatti,

"leggiadri"/graceful preceded by "studi"/studies followed by "talor"/sometimes with final apocope
SUBWORD UNITS = i, ò, erò
studii, studiò, studierò // FULL--- di, i, la, le, in, ',', il

"attende"/awaits preceded by clitic pronoun "La"/her followed by "a"/in
SUBWORD UNITS = ssù, scinado, sci, vi, vata, scio, vando, vata // BUT--- Lassù, Lasciando, Lasci, Lavi, Lavati, Lascio, Lavando, Lavata, FULL--- mattina, porta

"aguzzi"/spiky preceded by "cocci"/shards followed by "di"/of
SUBWORD UNITS = uoli, e, i, umi, u, uti, uto // cocciuoli, coccie, coccii, cocciumi, cocciu, BUT--- cocciuti, cocciuto, FULL--- pieni, grossi, duri

"silenzio"/silence preceded by "pian"/plane with final "e" elided followed by "verde."/green.
SUBWORD UNITS = ero, etto, tone, tato, tare, ure, ura, gente, issimi // pianero, pianetto, BUT--- piantone, piantato, piantare, pianure, pianura, piangente, pianissimo, FULL--- del,

From the data reported above, it can be easily concluded that BERT is totally unable to produce a prediction in two positions: sentence start and sentence end. Then, if we consider case c., we can assume that besides the objective fact that the masked word is not present in the dictionary of the model, there is another important factor contributing the wrong outcome: the lack of a meaningful word following the masked word which obliges BERT to concentrate the attention to the previous one.

Coming to case d., we can see that the presence of meaningful words following the masked word contributes some context that produces a better outcome, generating substituting words of the same grammatical class of the masked one: if the word is a noun, it produces substitutes which are nouns, the same if the word is a verb.

Case 'e' is more complex. BERT is unable to produce substitutions and can only generate subword units. The context is insufficient or it is highly unpredictable. This is true for the three cases in which the masked word is not followed by punctuation: "esili"/slender, "come"/as, "eseguire"/accomplish. The three sequences are as follows:

"slender thoughts", "as of autumn", "the accomplishing deeds"

Case 'f' being the most frequent is the one with the highest level of predictability compared to the previous cases. In fact, most masked words are followed by a conjunction "e" of a preposition. Those cases in which the masked word is followed by a semantically important word are just strongly infrequent. Let's consider them one by one:

"sovrumani"/superhuman followed by "silenzi"/silences
"silenzio"/silence followed by "verde"/green
"leggiadri"/graceful followed by "talor"/sometimes with final apocope
"conversare"/talking followed by "l'anima"/the soul

So we come up with the following four sequences: "superhuman silences", "silence green", "graceful sometimes" and "talking the soul", where the last two cases are due to discontinuities in the overall syntactic structure of the sentence.

## 6. The Second Phase: Comparing Contextual with Frequency Values

In the second phase we repeated an experiment already reported in previous papers (see [27-28]), this time searching 10 candidates rather than simply five. In order to evaluate the CONTEXT from a syntagmatic and paradigmatic point of view, we duplicated the marked sentences imposing a canonical structure, thus partially deleting the surprise effect caused locally by adjacency of a head noun and a dislocated or just discontinuous linguistic item belonging to another constituent. In a section below are the description of the non-canonical features of the sentences we used for the experiment. The English translation is available in the Appendix. We signed every sentence with letter A for those belonging to the poetry domain - 7, and letter B for newswire domain - 11. The newswire sentences are taken from the treebank of Italian – VIT, Venice Italian Treebank – available also under UD repositories at https://universaldependencies.org; the poetry set of sentences is taken from publicly available collections of poets of the first half of the nineteenth century which have already undergone specific analysis in previous work(see [2;3]).

In Table 4. we evaluate the seven sentences from the poetry domain, and in Table 5. the eleven sentences from the newswire domain. We computed three main parameters: in column 2, Number of Words masked with respect to total number of tokens; in columns 3 and 4 we list words correctly predicted with the identical corresponding word respectively in the Non Canonical and in the Canonical sentence structure; then in columns 5 and 6 we list the number of words with frequency values respectively Higher and Lower than a given threshold that we established at 10.000 occurrences. We also considered words that don’t appear in the 50000 vocabulary and reported them after a slash: we assume their import should be valued double. Thus for instance, in the Poetry text, we found 5 such words and the total number of Low Frequency Words is thus increased by 10 points.

Table 4. Evaluation of Poetry Sentences

| Sent. No. | No. Words/Masked | Non Canon. Words | Canon. Words | High Freq. Words | Low Freq. Words |
|---|---|---|---|---|---|
| 2.A | 10/8 | 0 | 3 | 4 | 3/1 |
| 3.A | 14/9 | 3 | 4 | 6 | 3 |
| 4.A | 10/8 | 2 | 2 | 4 | 4 |
| 5.A | 9/6 | 0 | 0 | 4 | 1/2 |
| 12.A | 11/7 | 1 | 2 | 4 | 1 |
| 13.A | 15/7 | 0 | 0 | 5 | 0/2 |
| 14.A | 14/9 | 1 | 1 | 6 | 3/1 |
| totals | 83/54 | 7 | 12 | 33 | 15/6=27 |
| ratios | 0.65 | 0.583 | | | 0.818 |

Table 5. Evaluation of Newswire Sentences

| Sent. No. | No. Words/ Masked | Non Canon. Words | Canon. Words | High Freq. Words | Low Freq. Words |
|---|---|---|---|---|---|
| 1.B | 14/8 | 3 | 5 | 8 | 0 |
| 6.B | 6/5 | 2 | 3 | 5 | 0 |
| 7.B | 5/4 | 0 | 0 | 3 | 1 |
| 8.B | 10/7 | 1 | 2 | 6 | 1 |
| 9.B | 7/4 | 1 | 1 | 4 | 1 |
| 10.B | 12/9 | 1 | 1 | 7 | 2 |
| 11.B | 15/10 | 2 | 4 | 10 | 0 |
| 15.B | 25/10 | 7 | 7 | 8 | 2 |
| 16.B | 22/10 | 4 | 4 | 8 | 2 |
| 17.B | 15/9 | 6 | 6 | 10 | 0 |
| 18.B | 22/10 | 4 | 4 | 9 | 0/1 |
| totals | 153/86 | 30 | 36 | 78 | 9/1=11 |
| ratios | 0.56 | 0.834 | | | 0.141 |

As can be easily noticed by comparing all parameters, poetry and news have opposite values. Quantities measured in column 2 show how the ratio of masked words is higher in poetry than in the news domain – 0.65 vs 0.56 -, the reason being that poetry text makes use of less grammatical or function words, like articles, clitics, prepositions which are highly predictable but are less informative. The first important parameter is the difference in number of masked words identified in Non-Canonical vs Canonical Sentences, and here again as can be easily noticed the newswire domain has a much higher score than the poetry domain – 0.834 vs 0.583. The newswire domain has a very limited advantage in the conversion from non-canonical to canonical, the difference being less than 17 point. In the poetry domain, on the contrary the difference overrides 40 points, being thus highly significant.

Then the second relevant parameter derived by the proportion of High Frequency words vs Low Frequency words and computed as a ratio between the sum of the absolute number of words plus a doubling of the number of very low frequency words. Here the scores show the opposite relation, Poetry domain has a much higher number of Low Frequency words than Newswire domain – 0.818 vs 0.141.

The conclusion we can safely draw from these data is that in general the News domain has a higher linguistically and frequency-based evaluated prediction score:

- ✔ because it has a much lower number of Low Frequency words

✔ because it has a higher number of contextually predictable words in Non-canonical structures

In other words, the relevance of context varies according to the domain: in the Poetry domain it is both dependent on word frequency and context, i.e. word structural position, but context seems more relevant. Not so in the Newswire domain where context varies less and frequency plays a higher role.

One example is highly representative of the interplay between frequency and context and is the word "Ora", an ambiguous word with two homographs-homophones: one meaning "now", an adverbial contained in sentence n. 9 – of the newswire domain; and another meaning "hour", a (temporal) noun, contained in sentence n. 5 – of the poetry domain. Only the adverbial is predicted in both structural versions. On the contrary, the noun is contained in a sentence belonging to the poetry domain where the overall context is not supportive for that word predictability.

Below, we list the words which have been assigned a cosine value higher than 0.5 in canonical and non-canonical structures. All cases of non-canonical structures are included in canonical ones where four additional words are present.

Table 6. Best cosine values for identically predicted masked words

| Sent. No. | Masked Word | Cosine Value Non-Can. | Cosine Value Can. | Phrase Including | Lexical Type |
|---|---|---|---|---|---|
| 1 | miei | 0.88233 | | miei colleghi | Function |
| 1 | più | | 0.55960 | più acuta | Function |
| 11 | questo | | 0.76715 | questo libro | Function |
| 11 | esempi | 0.65383 | 0.73481 | esempi di carità | Content |
| 15 | come | 0.9186 | | come già | Function |
| 15 | ha | 0.97755 | | ha voluto | Function |
| 16 | viene | 0.79483 | | viene interrogato | Function |
| 16 | senatore | 0.80796 | | senatore a vita | Content |
| 16 | vita | 0.99582 | | senatore a vita | Content |
| 17 | fare | 0.81857 | | intervento da fare | Content |
| 17 | questi | 0.96136 | | questi giorni | Function |
| 17 | giorni | 0.83000 | | questi giorni | Content |
| 17 | detto | | 0.55038 | ha detto | Content |
| 18 | modo | 0.79384 | | modo di | Content |

As a general remark, in the comparison of function and content words we see in Table 6., function words have a much higher cosine score than content words – with the exception of the collocation or polirematic form: “senatore a vita”/life_long_senator, where both "senatore"/senator and “vita”/life receive a high cosine value, again confirming the relevance of the context, which in this case is as relevant as that one of function words and is the most important parameter to consider.

In sum, in the poetry domain the two main parameters – word frequency and word context - conspire to reduce the predictability score. The context in poetry domain is characterized by metaphorical usage of word combination thus dramatically reducing the ability of BERT to find embeddings. Thus context has a double dimension: word combination aiming at producing metaphorical meaning is accompanied by constituent displacement and discontinuity contributing surprisal effects due to non-canonical structures. These two aspects are further constrained by the low frequency of some words thus justifying the low value of the overall predictability parameter. The opposite applies to the news domain: word linear combinations remain fairly literal in their semantic usage thus favouring the possibility for BERT to find embeddings even when words have low frequency values. Absolute frequency is thus less relevant in the Newswire than in the Poetry domain.

## 7. Conclusions and Future Work

In this paper we have proposed a number of experiments based on BERT's first layer, the so-called raw embeddings, that we have tested in a highly constrained context with a corpus of 80 sentences extracted from Italian poetry text of the beginning of last century. The challenging context was characterized by the combination of three fundamental factors for a sentence meaning understanding perspective on the prediction task represented by BERT masked task: use of infrequent words - as measured against the ItWac frequency list - and their phrase level combination – word poetic usage for metaphors w.r.t. possible semantic association -, and their larger sentential context in uncommon syntactic structures – non-canonical structures. In order to be able to evaluate the different impact of the three adversarial factors on masked word prediction, we have included in the dataset a small number of sentences from newswire domain showing the same structural syntactic properties but lacking both the usage of very infrequent words – with a few exceptions - and their uncommon combination to produce metaphors.

The results have clearly shown the ability of newswire sentences to receive an overall higher word predictability score thanks to the smaller effect of adversarial factors we investigated. The answer to the question: is frequency or context the determining factor for Transformer Language Models to predict the masked word, is both are, but their relevance depends on the domain. The newswire domain has less infrequent words and less uncommon non-canonical structures than the poetry domain, which is what explains the remarkable difference in final results. On the opposite, the poetry domain has very infrequent words and highly uncommon non-canonical structures which makes it very hard to predict. Improvements from the conversion of noncanonical to canonical structures are very high and significant thus showing that CONTEXT is more relevant than FREQUENCIES in the evaluation of word predictability.

In future work we intend to use sentences contained in the treebank of Italian called VIT [3] - which is made up of 11,000 sentences - where some 30% of sentences have been manually classified as non-canonical. Using this dataset we will produce a set of experiments: in a first experiment we will fine-tunea BERT model for Italian with VIT corpus, and then we will use the 51 poetry sentences – or a larger text - as test set. In another experiment always based on Machine Learning, we will be using some variant of BERT at first with a totally unsupervised approach, and finally a fully supervised approach also introducing syntactic information as has been done in a recent task we participated [29]. This will be done in order to verify the powerful ability of Transformers as classifiers and test it with a binary task: canonical vs non-canonical sentences.

## APPENDIX - ENGLISH VERSION OF THE CANONICAL AND NON-CANONICAL TEXT

1.B Today I thank for the courtesy on several occasions demonstrated to me and my colleagues. 2.A She alone maybe the cold dreamer would educate to the tender prodigy. 3.A I think of a green garden where with you resume can conversing the soul maiden. 4.A If spring my generous heart choked of deaf spasms. 5.A Neither the oblivious enchantment of the hour the iron-like beat grants. 6.B Becomes thus sharper the contradiction. 7.B Good instead overall the rest. 8.B An important decision Ghitti reserved after the holidays. 9.B The important thing is now to open it more. 10.B His information would also give to the guidelines of laique democracy greater boosts. 11.B In this book Maria Teresa, they explain at Mondadori's, will give examples of charities concrete. 12.A Said that they have his heart from inside the chest removed. 13.A The reluctant opinions and not ready and in the midst of executing works hampered. 14.A An echo of mature anguish revverdived to touch signs to the flesh dark of joy. 15.B The government, therefore, though giving up the absolute majority, has wanted, as already in IMI, focusing on a gradual privatization. 16.B At a conference in the Viminale the minister, when he is questioned on the senator to life, at first does not understand the name. 17.B First intervention to do, he said these days, is to implement the reform. 18.B I conceive the private as a work method, as work contracts, as a way to manage in short.

1.Bc Today I thank you for the courtesy demonstrated to me and my colleagues on several occasions. 2.Ac Maybe the cold dreamer educated her alone to the tender prodigy. 3.Ac I think of a green garden where the soul maid can resume conversing with you. 4.Ac Spring if you choked my generous heart of deaf spasms. 5.Ac Neither the iron-like beat of the hour grants the oblivious enchantment. 6.Bc The contradiction becomes thus sharper. 7.Bc Instead, overall the rest is good. 8.Bc Ghitti reserved an important decision after the holidays. 9.Bc Now it's important to open it more. 10.Bc His information would also give greater boosts to the guidelines of laique democracy. 11.Bc In this book Maria Teresa will give concrete examples of charities, they explain at Mondadori's. 12.Ac They said they took off his heart from the chest. 13.Ac The reluctant opinions and not ready works hampered in the middle of executing. 14.Ac An echo of mature anguish revverdressed to touch signs of joy obscure to the flesh. 15.Bc So the government wanted to focus on a gradual privatization while giving up the absolute majority as already in IMI. 16.Bc At a conference in the Viminale, when he is questioned on the senator to life at first the minister does not understand the name. 17.Bc To implement the reform is first intervention to do, he said these days. 18.Bc I conceive the private as a work method, such as work contracts, as a way to manage in short.

## SUPPLEMENTAL MATERIAL

**Sentence 1.B -** Oggi ringrazio della cortesia in più occasioni dimostrata a me e ai miei colleghi. 1.Bc Oggi ringrazio della cortesia dimostrata a me e ai miei colleghi in più occasioni.

The sentence belongs to the newswire domain: it is computed best in the canonical form, with 5 words over 8 while the non-canonical version has only 3 words predicted correctly – only ”più/more”, "occasioni/chances" and "miei/my". Cosine values are not particularly high except for "miei/my" the possessive which being in its attributive position has a favourable predictive condition. “Oggi” is wrongly predicted as being a separator with very high value, “ ‹s› 0.99998”. It can be noted that “ringrazio” is partially predicted by “Grazie” in first position but very low value 0.14397. Now the canonical version:

Ringrazio (0.0238), più (0.287), occasioni (0.545), dimostrata (0.165), miei (0.882). Interesting to note that the three words predicted in both structural versions have the same cosine values. When we add the remaining 7 sentences, another word is predicted, colleghi (0.076). No connection with frequency values of the missing words: they are all positioned in the high part of the frequency list – excluding "più" and "miei" which are grammatical words and are positioned close to the top.

Frequency List: °-più; °-miei; °-Oggi; °-colleghi; °-occasioni; °-ringrazio; °-dimostrata; °-cortesia

**Sentence 2.A -** Lei sola forse il freddo sognatore educherebbe al tenero prodigio. 2.Ac Forse il freddo sognatore educherebbe lei sola al tenero prodigio.

The second sentence belongs to the poetry domain. The original non-canonical version has no candidate found in the first 5 positions. This may be due to presence of a rather infrequent word like "educherebbe/would+educate" as main verb which only appears listed low only in the Upper List. On the contrary, the canonical form has three words predicted: first "Forse/Maybe ", second word "lei/She", and third word "solo"/alone but with wrong masculine morphology. However, these words are correctly predicted with low cosine values - Forse (0.149), lei (0.0355) solo (0.0145). No version provides useful approximations of the meaning of the missing words even though "freddo/cold" is included in the high portion of the 50000 vocabulary. As to the remaining words, they are still included in the Vocabulary but in the lower portion. It is important to note that the lack of prediction can only be motivated just because by combining not so frequent words in unusual combination has produced metaphors like "cold dreamer", "tender prodigy", in association with a verb like "educate".

Frequency List: °-solo; °-lei; °-Forse; °-freddo; *-tenero; *-prodigio; *-sognatore; **-educherebbe

**Sentence 3.A -** Penso a un verde giardino ove con te riprendere può a conversare l'anima fanciulla. 3.Ac Penso a un verde giardino ove l'anima fanciulla può riprendere a conversare con te.

The non-canonical version of this sentence has two words correctly predicted, giardino/garden, ove/where and a third word with different morphology, in slot 5, Pensa/Think(3rd+person+singular+present+indicative), rather than Penso(1st+person). In the canonical version we find correctly Penso/think in second slot, and another word is added può/can, the modal auxiliary that is now positioned correctly in front of its main verb "riprendere/restart", which is by itself a very frequent verb. As to cosine values, we have the following low values for the canonical version: Penso (0.085), giardino (0.194), ove (0.146), può (0.0865). The non-canonical version has a lower value for Penso but a higher value for giardino (0.291). In the longer context, the interesting fact is constituted by the substitution of "Pensa" with fino/until in the non-canonical version; while in the canonical version Penso/think is moved to a worse position from second slot to last slot, slot 5 and a lower cosine value (0.06112). As to the non-predicted noun modifier "fanciulla/maid", this is certainly an unusual combination even though the two words are highly frequent. The result of the combination is of course a beautiful metaphor which combines "primavera"/spring with "fanciulla"/maid and the garden. Notice the different position of Penso+1st+pers, with respect to Pensa+3rd+pers which is by far less frequent. Now consider the word conversare/conversing which receives the following list of non-word predicted candidates: erare/?? (0.4455), rare/rare?? (0.16737), lare/?? (0.0549), mare/sea?? (0.0479), scere/?? (0.03124). Apart from RARE and MARE which I don't regard being selected for their current meaning but just for being part of the list of subwords, the remaining segments are all meaningless and bear no semantically useful relation with the masked word CONVERSARE.

Frequency List: °-può; °-ove; °-anima; °-verde; °-Penso; °-riprendere; *-Pensa; *-fanciulla; *-conversare

**Sentence 4.A -** Se primavera il mio cuor generoso soffocasti di spasimi sordi. 4.Ac Primavera, se soffocasti il mio cuor generoso di spasimi sordi.

In this sentence only the phrase "mio cuor"/my heart is predicted in both structural versions. mio (0.291), cuor (0.394). The word "Primavera", which is the first word in the canonical version, has no close prediction: as happens in all sentences, the prediction is totally missed whenever a content word appears in first position. In the non-canonical version, the word comes second, after the conjunction "Se"/If, which predicts the appearance of an auxiliary BE/HAVE in their correct morphological word form – fossi/were, avessi/had in both cases with first person morphology, but also fosse/were, and the last two: con/with and solo/alone. The version with the addition of the 7 sentences has the worsening effect of introducing a subword in place of con/with, MMAI which I assume derives from the wrongly split SEMMAI/if+ever. The word has been wrongly split because the segment SE is wrongly – at least in the word SEMMAI - regarded as a legitimate segment due to its very high frequency. Again the problem seems the unusual combination of the remaining words which are fairly common, apart from soffocasti/choked which is not included in the frequent nor in the Rare wordform list; and spasmi/spasms which is only included in the Upper List. In other words, it's their metaphorical import that prevents the correct prediction. However, it is the position that produces the worst results: the adjective "sordi/deaf" in predicative position is predicted as a punctuation mark in both structural versions.

Frequency List: °-Se; °-mio; °-cuore; °-primavera; *-generoso; *-Primavera; *-sordi; **-spasmi

**Sentence 5.A -** Né l'oblioso incanto dell'ora il ferreo battito concede. 5.Ac Né il ferreo battito dell'ora concede l'oblioso incanto.

This sentence is the worst case of the poetry domain lot: it has no word predicted neither in the non-canonical nor in the canonical version. This may be due to the presence of a very infrequent word "obliosi/oblivious". However, we notice the presence of an unusual combination of the attributive metaphoric use of "ferreo/iron-like", a rather unusual word. But of course, it is just the combination of words used to build a powerful metaphor that prevents predictions to take place. It is worthwhile noting that "incanto"/enchantment is substituted by ten candidates semantically loosely related to the domains evoked by the masked word: temporal dimension (rhythm, stepping, passing, proceeding, beat), and a condition of the contemplating mind (silence, rest, meaning, thought, sound). Also another important remark regards the inability to predict the ambiguous word "ora"/hour, homograph with "ora"/now, thus clearly showing that context is the determining factor.

Frequency List: °-ora; °-Né; °-concede; °-incanto; *-battito; **-ferreo; **-oblioso

**Sentence 6.B -** *Diventa* così più acuta la contraddizione. 6.Bc La contraddizione diventa così più acuta.

This sentence has different predicted words in the two structural representations, Diventa/Becomes is present in both. Then "così/so" and "più/more" are predicted in the canonical sentence - diventa (0.215), così (0.0439), più (0.559); while in the non-canonical structure only acuta/sharp is predicted, acuta (0.0441), and the cosine value for "Diventa" is lower being in sentence first position. The canonical form has predicted the discourse marker "così/so" positioned in sentence center: not so in the non-canonical structure where we can again assume that it is the position right after the verb at the beginning of the sentence that does not allow the prediction, notwithstanding its high frequency. Now consider the high frequency of "contraddizione" which is not predicted presumably because of its position at the end of the sentence: the first candidate is the subword "mente" with cosine value (0.16536), followed by sensibilità/sensibility, coscienza/conscience, gioia/joy.

Frequency List: °-più; °-così; °-contraddizione; °-acuta; *-Diventa

**Sentence 7.B -** Buono invece in complesso il resto. 7.Bc Invece in complesso il resto è buono.

No word was predicted in either versions. In order to transform the original non-canonical version in the corresponding canonical one we added the copula "è" that is missing in the original sentence. This is predicted in the canonical version but since it has been added we do not count it for the actual predictive task. All the words are very frequent. As will be clarified further on, whenever the first word of the sentence coincides with a discourse marker or a conjunction the prediction is very close if not equal. This is the case for the canonical form of the sentence starting with "Invece"/Rather, which has the five following best predictions: "Ma"/But, "E"/And, "Però"/However, "Più"/More, "Ed"/And, all belonging to the same grammatical category and in two cases, also to the same semantic type ("Ma", "Però"). Considering the status of the adjective "Buono"/Good which comes in first position in the non-canonical structure and in second position in the canonical one, one can clearly realize the importance of the respective position and the context on the ability of BERT to predict. In the first case, the word coming first position has no left context and there is no similarity, not even at a grammatical level: only conjunctions and verbs are predicted. On the contrary, in the canonical form, “buono” appears as predicate in a copulative structure and the predictions are very close: diverso/different, risolto/resolved, compiuto/achieved, secondario/secondary, positivo/positive.

Frequency List: °-invece; °-resto; °-complesso; *-Buono

**Sentence 8.B -** Una decisione importante Ghitti l'ha riservata a dopo le feste. 8.Bc Ghitti ha riservato una decisione importante a dopo le feste.

Only one word is predicted in both versions but it is not the same word. The canonical version predicts "importante/important", (0,0605), the non-canonical version predicts "dopo/after", (0.0152). As can be noticed, the cosine values are very low and again the frequency of occurrence of the words contained in the sentence is fairly high - excluding the proper name “Ghitti” which does not exist in the overall frequency list. The unexpected fact is constituted by the inability to predict the auxiliary “ha”/has in the non-canonical structure – as opposed to what happens in the canonical one -, and the association in fourth slot of a non-word like “vamteen“, presumably a subword of some kind. The only explanation could be the presence of a past participle with feminine+singular ending which is only allowed by presence of the resumptive clitic “la” needed to construct the Clitic Left Dislocation of the object NP “Una decisione importante”. As said above, the canonical version predicts the presence of the auxiliary HAVE in the correct form and also in two additional morphologically possible forms: “aveva”/had+3rd+pers and “avrebbe”/would+have+3rd+pers; final word predicted in the other auxiliary legal form “è”/is.

Frequency List: °-dopo; °-importante; °-decisione; °-riservata; °-feste; ***‹ukn›-Ghitti

**Sentence 9.B -** L'importante ora è aprirlo di più. 9.Bc Ora è importante aprirlo di più.

This sentence is perhaps too short and only function words are captured by BERT embeddings: ora/now (0.3825) più/more (0.0911). The ambiguous word "ora"/now is better predicted in the non-canonical structure - in first position - for the availability of right context - the canonical version predicts "Ora" in fourth position (0.0844). Again this is not relatable to a frequency problem but just structural problems, with the exception perhaps of the final word "aprirlo" which is only present in the very-low frequency list. In fact, in the canonical version, "aprirlo"/open+it is substituted by cliticized verbs - though semantically unrelated, however, showing that the morphology has been captured correctly. As to "importante"/important, it does not appear in the first five candidates, but it is predicted in sixth position (0.04902).

Frequency List: °-ora; **-aprirlo

**Sentence 10.B -** Le sue informazioni darebbero anche agli orientamenti di democrazia laica maggiori spinte. 10.Bc Le sue informazioni darebbero maggiori spinte anche agli orientamenti di democrazia laica.

This sentence has the same predicted word "maggiori/major" in both structural representations. As before, the words are all very frequent with the exception of “darebbero/+would+give, which is below the threshold and is only part of the “very+low” List. Now consider the word spinte/boosts: predicted masked words are as follows: certezze/certainties (0.0852), garanzie/guarantees (0.0824), informazioni/information (0.04183), taria/tary (0.04003), opportunità/opportunities (0.0383). The fourth slot contains a subword, in fact a non-word, which is assigned a score higher than the one assigned to “opportunities”. The question is that the masked word is not frequent enough to be able to collect the co-occurrences required. As a result, even very low scored embeddings are considered. The non-word gets a slightly better score when the text is considered as a whole with the last 7 sentences added, up to (0.06002), but remains always in fourth position.

Frequency List: °-anche; °-informazioni; °-sue; °-maggiori; °-democrazia; °-orientamenti; °-laica; *-spinte; *-darebbero

**Sentence 11.B -** In questo libro Maria Teresa, spiegano alla Mondadori, darà esempi di carità concreti. 11.Bc In questo libro Maria Teresa darà esempi di carità concreti, spiegano alla Mondadori.

In this sentence there is a striking difference in prediction between the two structures. The non-canonical version has only two words predicted, "libro/book" and "esempi/examples", libro (0.0242), esempi (0.653). On the contrary, in the canonical version BERT manages to predict four words, "questo/this", "Maria/Mary", "Teresa/Therese", "esempi/examples", questo (0.767), Maria (0.283), Teresa (0.141), esempi (0.734). Strangely enough, the word "libro" does not figure in the first five candidates. Useless to say, the remaining words are all very frequent. The third run with a longer text including the following 7 sentences gives interesting results: "Teresa" now becomes first candidate substituting the previously chosen first candidate "ci"/us. The word "esempi"/examples, predicted as first candidate, in the text is followed by "carità"/charity which is not predicted in both version: in its place, the first candidate is again "esempi", thus certifying that predictions are made one word at a time disregarding the textual context. Now consider the adjective "concreti" which has been dislocated and is disjoined from its head, "esempi". The list of five candidates for the canonical version is the following: "cristiana+fem+sing"/Christian (0.1919), '.' (0.0909), ',' (0.0387), "civile+sing"/civil (0.0383), "esemplare+sing"/exemplar (0.0222). None of the candidates is plural in number as it should be, if the morphology of Italian has to be respected. On the contrary, the first candidate agrees both in number and gender with the preceding word "carità+fem+sing"/charity, which is not to be considered the correct nominal head. The non-canonical version has one punctuation mark less and an additional adjective "pastorale+sing"/pastoral. \
Frequency List: °-questo; °-libro; °-esempi; °-carità; °-concreti; °-darà; °-spiegano; °-Mondadori

**Sentence 12.A -** Disse che gli hanno il cor di mezzo il petto tolto. 12.Ac Disse che gli hanno tolto il cuore di mezzo il petto.

This sentence from the poetry subset has only one word in common "cor/heart" and an additional word predicted in the canonical structure, "tolto/taken+off". The cosine values are all very low, cor-cuore (0.1019), for the non-canonical, and cor-cuore (0.0756), tolto (0.156) in the other structure. Interesting enough, when using the configuration with the whole text, also "mezzo/means" is predicted in second slot.
Frequency List: °-mezzo; °-cuore; °-petto; °-tolto; *-Disse

**Sentence 13.A -** I ritrosi pareri e le non pronte e in mezzo a l'eseguire opere impedite. 13.Ac I ritrosi pareri e le opere non pronte e impedite in mezzo a l'eseguire.

No prediction found by BERT in the two structural representations - with the exception of "mezzo"/means which however is only appearing in 8th position and not considered in this evlauation. However it is important to note that the previous seven predicted words are in fact only subwords, mostly meaningless, and some having a corresponding identical wordform with a totally different meaning. Here they are: "dotti"/learned+mas+plur, "dotte"/learned+fem+plur, "tente"/meaningless, "sistenti"/meaningless, "sistenza"/meaningless, "difficoltà"/difficulty, "fami"/meaningless. As to their frequency, words are mostly frequent but there are two missing words in the overall frequency lists: "ritrosi/reluctant" and "impedite/hampered". These two words may have been supplemented as subwords but with no useful context for the current analysis. The five candidates appearing are as follows: for "ritrosi" we have - suoi/his+hers, non/not, buoni/good+masc+plur, mal/bad(truncated), loro/their+them+they; and for "impedite" - '.', buone/good+fem+plur, inutili/useless+plur, nuove/new+fem+plur, pubbliche/public+fem+plur. In all of these cases, even if the correct word has not been predicted, the morphology has been matched correctly.

Frequency List: °-mezzo; °-opere; °-pareri; °-eseguire; °-pronte; ***ritrosi; ***impedite

**Sentence 14.A -** Un'eco di mature angosce rinverdiva a toccar segni alla carne oscuri di gioia. 14.Ac Un'eco di mature angosce rinverdiva a toccar segni di gioia oscuri alla carne.

This is another sentence from poetry domain very hard to tackle and to understand. Both the canonical and the non-canonical analyses have just one word found, "eco/echo" (0.0984). Of course the main verb "rinverdiva" is not amongst the frequent words in the list: in fact, it is missing. The remaining words are frequent but they are organized in a peculiar structural configuration with the declared aim to produce metaphors. No changes or improvements when the sentence is analysed with the canonical version of the text. As we did for example 11, we now consider the discontinuous adjective “oscuri+masc+plur”/obscure and the morphology of the five candidates predicted. In the non-canonical version we have: “pieni+mas+plur”/full (0.5461), “piena+fem+sing”/full (0.0486), “e”/and, ‘,’, “pieno+mas+sing”/full (0.0216). Now the canonical version: “fino”/until (0.1139), “intorno”/around (0.1139), “dentro”/inside (0.1001), “sino”/until (0.0476), “vicino”/close (0.0437). As can be noticed, all of the predicted words for the non-canonical structure are function words and none – with the possible exclusion of the ambiguou “vicino+mas+sing” - is an adjective. The reason for this lack of grammatical match may be due to the presence of the articulated preposition “alle”/to the+fem+plur in the canonical version. In the non-canonical version the word “oscuri” was followed by a preposition “di” which is the most frequent wordform with 65 million occurrences.

Frequency List: °-alla; °-carne; °-gioia; °-segni; °-toccare; °-eco; *-oscuri; *-mature; *-angosce; ***rinverdiva

**Sentence 15.B -** Il governo, quindi, pur rinunciando alla maggioranza assoluta, ha voluto, come già nell'IMI, puntare a una privatizzazione graduale. 15.Bc Quindi, il governo ha voluto puntare a una privatizzazione graduale pur rinunciando alla maggioranza assoluta come già nell'IMI.

This long sentence belongs to the domain of the news and even in its non-canonical structure, it is more linear and thus more predictable. There are seven words predicted (over ten we masked) in the two versions: governo/government (0.304), maggioranza/majority (0.0377), assoluta/absolute (0.349), ha/has (0.977), voluto/wanted (0.491), puntare/aim (0.0385). The proper name IMI is in the very low list. Strangely enough the function word come/like (0.1925/0.9186) is predicted as first candidate in its non-canonical position, as second position ,but with a much lower cosine measure in canonical position.

Frequency List: °-governo; °-maggioranza; °-voluto; °-assoluta; °-puntare, °-privatizzazione; °-graduale; *-rinunciando; **-IMI

**Sentence 16.B -** In una conferenza al Viminale il ministro, quando viene interrogato sul senatore a vita, sulle prime non capisce il nome. 16.Bc In una conferenza al Viminale, quando viene interrogato sul senatore a vita sulle prime il ministro non capisce il nome.

There are four words predicted in this long sentence, again in the domain of the news, in the canonical and the non-canonical structures. They are: ministro/minister (0.497), viene (0.795), senatore/senator (0.808), vita/life (0.996). Again, most words are very frequent. An apparent difficulty is constituted by presence of a multiword: "sulle prime/at first" which may be hard to distinguish and differentiate on the basis of the context. In fact, in both structures, “prime” is substituted by riforme/reforms, banche/banks, dimissioni/resignation , pensioni/pensions, cose/things.

Frequency List: °-vita; °-viene: °-nome; °-ministro; °-prime; °-senatore: °-conferenza; °-capisce; *-interrogato; *-Viminale

**Sentence 17.B -** Primo intervento da fare, ha detto in questi giorni, è di attuare la riforma. 17.Bc Primo intervento da fare è di attuare la riforma, ha detto in questi giorni.

This is another fairly simple sentence which has the major number of predicted words in the whole set in relation to the total number in the sentence. There are six words predicted both in the canonical and the non-canonical version: "fare/do" (0.818), "ha/has" (0.283), questi/these (0.961), giorni/days (0.83), riforma/reform (0.194). The only difference being the slot assigned to riforma/reform, which has first slot in the canonical version and second slot in the non-canonical one, preceded by Costituzione/Constitution. Useless to say, the missing words are all very frequent.

Frequency List: °-fare; °-giorni; °-detto; °-intervento; °-riforma; °-Primo; °-attuare

**Sentence 18.B -** Io il privato lo concepisco come un metodo di lavoro, come contratti di lavoro, come modo di gestire insomma. 18.Bc Io concepisco il privato come un metodo di lavoro, come contratti di lavoro, come modo di gestire insomma.

In this final sentence again belonging to the newswire domain, there are four words predicted: metodo/method (0.0618), lavoro/work (0.214), lavoro/work (0.214), modo/way (0.794). Again very frequent missing words, apart from "concepisco/surmise" which is the only word present in the Rare-Words list. When analyzed with the canonical version of the text, the word lavoro/work moves from third to first slot, with a slightly improved cosine score.

Frequency List: °-lavoro; °-modo; °-Io; °-contratti; °-privato; °-metodo; °-insomma; °-gestire; ***-concepisco.